\documentclass{bmvc2k}

\usepackage{algorithm}
\usepackage{algpseudocode}

\title{Exploiting Features \\with Split-and-Share Module}

\addauthor{Jaemin Lee}{j911.public@gmail.con}{1}
\addauthor{Minseok Seo}{msseok96@gmail.com}{1}
\addauthor{Jongchan Park}{jcpark@lunit.io}{2}
\addauthor{Dong-Geol Choi}{dgchoi@hanbat.ac.kr}{1}

\addinstitution{
 Hanbat National University, \\
 Daejeon, Korea
}
\addinstitution{
 Lunit Inc.\\
 Seoul, Korea
}

\runninghead{LEE, SEO, PARK, CHOI}{Exploiting Features with Split-and-Share Module}

\newcommand{\figref}[1]{Fig.~\ref{#1}}
\newcommand{\tabref}[1]{Table~\ref{#1}}

\newcommand{\sref}[1]{Sec.~\ref{#1}}
\newcommand{\jc}[1]{\textcolor{black}{#1}}

\begin{document}

\maketitle

\begin{abstract}

Deep convolutional neural networks (CNNs) have shown state-of-the-art performances in various computer vision tasks.
Advances on CNN architectures have focused mainly on designing convolutional blocks of the feature extractors, \jc{but} less on the classifiers that exploit extracted features.
\jc{In this work, we propose Split-and-Share Module (SSM), a classifier that splits a given feature into parts, which are partially shared by multiple sub-classifiers.
Our intuition is that the more the features are shared, the more common they will become, and SSM can encourage such structural characteristics in the split features.}
SSM can be easily integrated into any architecture without bells and whistles.
We have extensively validated the efficacy of SSM on ImageNet-1K classification task, and SSM has shown consistent and significant improvements over baseline architectures.
\jc{In addition, we analyze the effect of SSM using the Grad-CAM visualization.}
\end{abstract}

\section{Introduction}
\label{sec:intro}
Deep convolutional neural networks (CNNs) achieve high performance in various computer vision tasks~\cite{deng2009imagenet, lin2014microsoft, cordts2016cityscapes, kuehne2011hmdb}. A general anatomy of CNN splits the architecture into two parts: a feature extractor and a classifier~\cite{bengio2013representation}. A feature extractor consists of conv-blocks which are made of normalization layers, convolutional layers, non-linear activations~\cite{nair2010rectified}, and pooling layers. To design CNN architecture is to find a good conv-block and stack it repetitively. ResNet~\cite{he2016deep} added identity-based skip connections to the Conv-block to enable stable training even when the Conv-block is repeatedly stacked deeply. In addition, the Xception~\cite{chollet2017xception} structure is a network structure developed from the Inception~\cite{szegedy2015going} structure. Xception utilizes Depth-wise-separable convolution using 1x1Conv to significantly lower the computation of the network and even improve its performance.
Accordingly, the recent trend on neural architecture search~\cite{zoph2018learning, real2019regularized, tan2019efficientnet} focuses on designing better conv-blocks in a data-driven way. While the classifier is also a crucial part of a CNN, less attention has been paid on designing better classifiers. In this study, we focus on designing a classifier that further exploits a given feature vector. To the best of our knowledge, most of the CNN architectures simply adopt single or multiple linear combinations as the classifier. 
\jc{In this work, we propose a novel classifier, named Split-and-Share Module (SSM). SSM divides the given feature into several groups of channels, and the groups are partially shared among sub-classifiers. Each group of channels has different degree of sharing; and our intuition is that the mostly shared group will contain general features, and vice versa. This feature split-and-share method can encourage the diversity of the features by structure, and thus the diversity of the sub-classifiers, leading to higher performances when ensembled.}

\begin{figure}
\begin{center}
    \includegraphics[width=12.5cm]{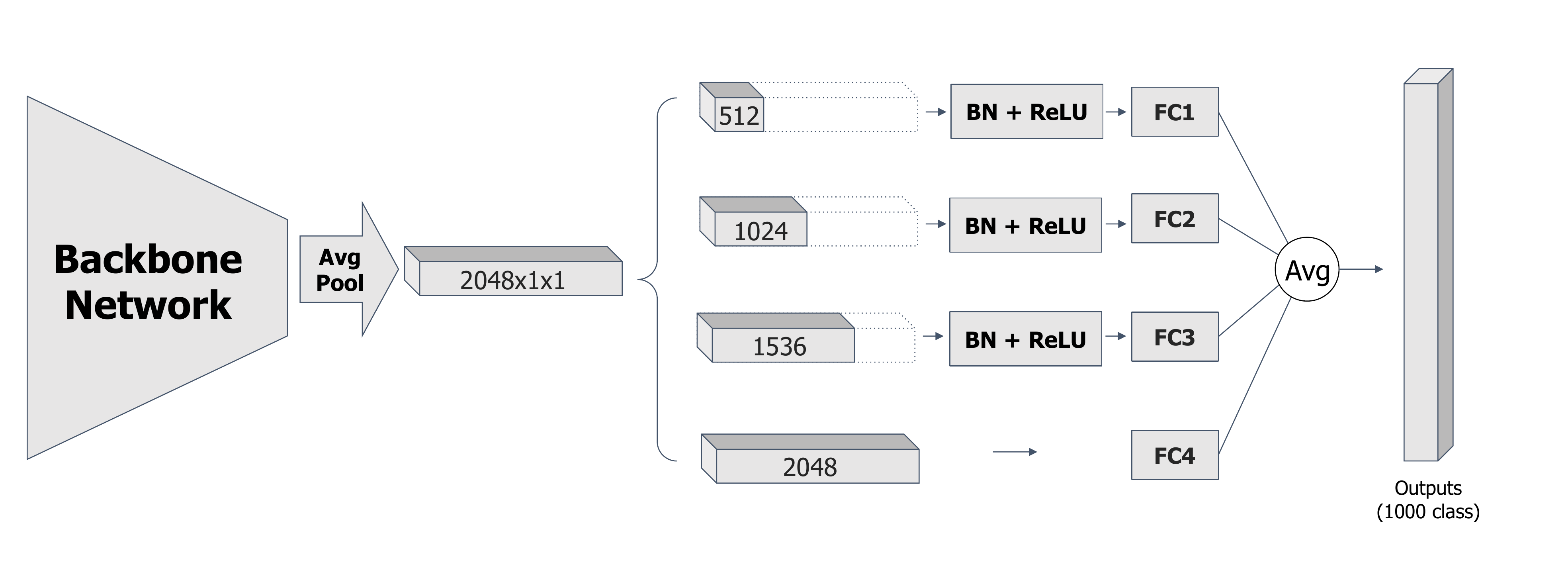}
    \caption{\textbf{An overview of SSM}. The illustrated example has 2048 channels in the final feature vector, and the output is 1000-way classification.}
    \label{figure:overview}
\end{center}

\end{figure}

\begin{figure}[!h]
\begin{center}

    \includegraphics[width=12.5cm]{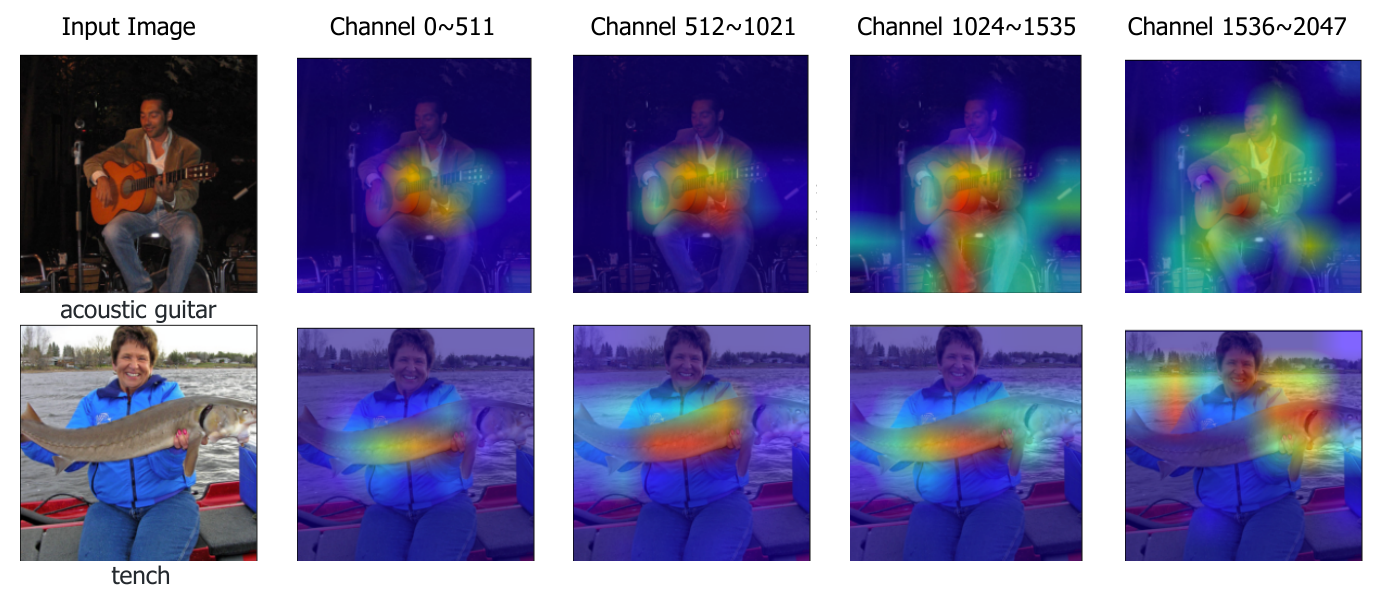}
    \caption{Grad-CAM visualization of channels with respect to sub-classifiers. ResNet-50+SSM is used for visualization, and the final feature size is 2048. Each column visualizes 512 channels. The first column is visualized with respect to FC1, the second column is visualized w.r.t. FC2, and so on. More details will be described in \sref{sec:qualitative_analysis}.}
    \label{figure:grad_cam1}
\end{center}
\end{figure}

Figure 1 shows the structure of the proposed SSM.
\jc{
Given a feature vector extracted from the backbone network (feature extractor), SSM splits the feature into 4 groups and each group is fed into the designated sub-classifier. The final output is the sum of outputs from each sub-classifier.
}
\jc{
The smallest group, illustrated as the bottom group in \figref{figure:grad_cam1}, is shared by all other sub-classifiers, and should contribute to the final prediction alone. It is encouraged to learn more common and general features in the limited number of channels. On the other hand, the least shared channels, illustrated as the top group in \figref{figure:grad_cam1}, will learn additional features such as contextual information.
}
\jc{
The Grad-CAM~\cite{selvaraju2017grad} visualization in \figref{figure:grad_cam1} qualitatively supports our intuition.
}
As shown in \figref{figure:grad_cam1}, the first column shows the acoustic guitar taken by Grad-CAM for each channel group. 
We can see that going down from the first row to the bottom row, starting with the additional characteristics of the acoustic guitar and gradually visualizing it as the core characteristic of the acoustic guitar. 
SSM shows stable performance improvement in architectures such as ResNet~\cite{he2016deep} and ResNeXt~\cite{xie2017aggregated}, and is a simple structure consisting of BatchNorm~\cite{ioffe2015batch} and ReLU, easy to attach to any CNN architecture.
\jc{
While the sub-classifiers may resemble the ensemble technique, which may lead to concerns on less improvements with ensemble. In our experiments, we show that a SSM-augmented network can further be improved with ensemble without any compromises.
}




\section{Related works}
\label{sec:relatedworks}

\subsection{Deeper architectures}

Starting with AlexNet~\cite{krizhevsky2012imagenet}, many CNN structures have been proposed. VGGNet~\cite{simonyan2014very} showed significant CNN performance improvement by increasing network depth. Another study made network learning stable by normalizing the input to each layer in batch units. Based on these developments, ResNet was proposed. ResNet proposed identity-based skip connections to deepen the network, greatly improving the performance of CNN. Since then, studies have been proposed to discover CNN structures through architecture search such as NASNet~\cite{zoph2018learning} and EfficientNet~\cite{tan2019efficientnet}. Also, architecture search methods based on Evolutionary Algorithms such as AmoebaNet~\cite{real2019regularized} have been proposed. Studies of these CNN structures have been continuously proposed and attracted great attention. But in the progress of the structure of these CNN classifier has been excluded. We conducted a study using features that were already well extracted from the feature extractor, and can be used in the architecture search later.

\subsection{Feature analysis}

Various analyses of features have been proposed. Ilyas et al.~\cite{ilyas2019adversarial} was experimental in that it exists as robust and non-robust features and can be classified, rather than simply dividing features into useful and useless features. From the perspective of Ilyas et al.~\cite{ilyas2019adversarial}, there are robust features and non-features among features used for prediction, both of which are useful for prediction but have different meanings and have room to utilize these characteristics.

Afalo et al.~\cite{aflalo2020knapsack} showed that using all the features that enter the classifier does not improve the accuracy of the CNN, but removing unnecessary features through pruning can improve the performance and the computational speed of the CNN.

In this respect, our SSM is also a new analysis and utilization of features. Our SSM forcibly assigns the role of features used for prediction through backprop, and qualitatively and quantitatively analyzes the effect on the network according to the location and number of such features.

\section{Split-and-Share Module}
\label{sec:fsm}

\jc{
In this section, we describe how SSM is formulated.
SSM is a simple classifier that splits and share features with multiple sub-classifiers.
The overall architecture of SSM is illustrated in \figref{figure:overview}, and the pseudo code algorithm is described in Algorithm~\ref{alg:fsm}.
}

\jc{
First, SSM equally divides the input feature in 4 splits, and sequentially append the splits one-by-one to formulate 4 features with different numbers of channels. For example, given the feature $F \in R^{2048}$, the first feature $F_1$ contains the first 1/4 channels, \textit{i.e.} $F_1=F[0:512]$. Accordingly, $F_2$ contains the first 1/2 channels, and so on.
In order to diversify the 4 features while keeping the feature domain with minimum overheads, we apply BatchNorm with ReLU to the first 3 features for simple scaling and non-linear activation. The resulting 4 features will have the same semantic meaning with different scales for the shared channels. Channels in the 4 features can be zeroed out by ReLU. BatchNorm and ReLU are essential in SSM, as they add extra non-linearity to the overall process.
Without BatchNorm and ReLU, SSM can be reduced to a simple linear combination (fully-connected) layer.
After splitting, recombining and re-scaling, the 4 features are feed-forwarded to 4 sub-classifiers. Each sub-classifier is a simple fully-connected layer, where the output dimension is the number of classes. 
The final output of SSM is the average of the 4 outputs from the sub-classifiers.
}

\jc{
The key intuition of our design is to partially share the given feature.
The first 1/4 channels are shared among all sub-classifiers. These channels are forwarded 4 times and back-propagated 4 times. As they are most frequently used channels, we expect these channels are trained to be the most important key features. In contrast, the last 1/4 channels are used only by the last sub-classifier, so they are expected to contain some additional features, such as context information on the surrounding environments.
We visualized the 4 splits of channels with the Grad-CAM visualization technique in \figref{figure:grad_cam1} and \figref{figure:grad_cam2}, and more analysis will be discussed in \sref{sec:qualitative_analysis}.
}

\newcommand{\factorial}{\ensuremath{\mbox{\sc Factorial}}}
\begin{algorithm}[h]
\caption{Split-and-Share Module}\label{alg:fsm}
\begin{algorithmic}[1]

\Procedure{SSM}{$features, num\_channels$=2048, $num\_heads$=4}

   \State $n\gets $INT$(num\_channels$ / $num\_heads$)
   \State $v \gets [(empty\_list)$]

   \For{$i$=1 to $num\_heads$}
      \State $out\gets features[:i * n]$
      \State $out\gets $BatchNorm$(out)$
      \State $out\gets $ReLU$(out)$
      \State $out\gets $FC$(out)$
      \State $v$.append($out$)
   \EndFor\label{fsmfor}
   
   \State $result \gets v$.sum() / $num\_heads$
       \State \textbf{return} $result$
\EndProcedure
\end{algorithmic}
\end{algorithm}

\section{Experiments}
\label{sec:experiments}

\jc{
In this section, we validate the efficacy of the proposed SSM on various architectures, and analyze the effect of SSM in several aspects.
First, we use SSM upon ResNet and ResNeXt architectures in ImageNet-1K classification dataset~\cite{deng2009imagenet}. SSM has shown performance improvements in most cases, and details will be described in \sref{sec:imagenet}. In \sref{sec:ablation}, we describe the ablation studies of SSM.
}


\subsection{ImageNet-1K classification}
\label{sec:imagenet}

The ImageNet-1K dataset~\cite{deng2009imagenet} consists of 1.28 million training images and 50k validation datasets. 
\jc{We follow the training details in ResNet~\cite{he2016deep}.
During training, the images are resized to 256x256 shape, and randomly cropped to 224x224 patches with random horizontal flipping.
During testing, the images are also resized to 256x256 shape, and a single 224x224 patch is cropped at the center.
For both training and testing, images are normalized with the mean and standard deviation of all pixels in the dataset.
We adopt He's method~\cite{he2015delving} for network random initialization.
We use SGD optimizer with base learning rate 0.1 and batch size of 256.
The running rate is reduced by one-tenth at epoch 30 and 60, and the total number of epochs is 90.}
The weight decay value is set to 0.0001 and the momentum value is set to 0.9. 

\begin{table}[h]
\centering
\resizebox{0.8\textwidth}{!}{%
\begin{tabular}{ l|c|c|c}
\hline
Architecture & Dataset & Epoch & Top-1 Acc \\
\hline
\hline
ResNet-18~\cite{he2016deep} & ImageNet-1K & 90 & 70.04\% \\
ResNet-18~\cite{he2016deep} + SSM & ImageNet-1K & 90 & \textbf{70.05\%}(+0.01\%) \\
ResNet-50~\cite{he2016deep} & ImageNet-1K & 90 & 75.65\% \\
ResNet-50~\cite{he2016deep} + SSM & ImageNet-1K & 90 & \textbf{76.68\%}(+1.03\%) \\
ResNet-101~\cite{he2016deep} & ImageNet-1K & 90 & 76.62\% \\
ResNet-101~\cite{he2016deep} + SSM & ImageNet-1K & 90 & \textbf{77.93\%}(+1.31\%) \\
\hline
ResNeXt50~\cite{xie2017aggregated} & ImageNet-1K & 90 & 77.19\% \\
ResNeXt50~\cite{xie2017aggregated} + SSM & ImageNet-1K & 90 & \textbf{77.96\%}(+0.77\%) \\
ResNeXt101~\cite{xie2017aggregated} & ImageNet-1K & 90 & 78.46\%\\
ResNeXt101~\cite{xie2017aggregated} + SSM & ImageNet-1K & 90 & \textbf{79.68\%}(+1.22\%) \\

\hline
\end{tabular}
}
\vspace{3mm}
\caption{\textbf{Classification results on ImageNet-1K.} Single-crop validation errors are reported.}
\label{table:imagenet}
\end{table}
 
\jc{
The experiment result is summarized in ~\tabref{table:imagenet}. SSM has consistently improved performance in all the architectures, except ResNet-18 that does not improve. 
The distinctive difference between ResNet-18 and other architectures is that the final feature of ResNet-18 has 512 channels, while others have 2048 channels.
Therefore, we assume that the number of channels in the final feature should be large enough for SSM to be effective.
}
In all architectures except ResNet-18, the performance improvement is significant.
\jc{
Furthermore, the absolute improvements in larger architectures are greater than the smaller ones. ResNet-101 improves 1.31\% in the top-1 accuracy, while ResNet-50 improves 1.03\%; ResNeXt-101 improves 1.22\%, while ResNeXt-50 improves 0.77\%.
}

\subsection{Ablation studies and analysis}
\label{sec:ablation}

\subsubsection{Training scheme for sub-classifiers}
\jc{There are 2 simple ways to train the 4 sub-classifiers: apply the classification loss to individual sub-classifier outputs, or apply the loss to the average of the outputs.
The former one requires each sub-classifier to independently learn to classify, and then ensemble the 4 sub-classifiers; the latter one allows the sub-classifiers to jointly learn to classify.
The results are summarized in \tabref{table:ablation1}. 
When individually trained, the sub-classifiers' performances are much higher than the jointly trained ones.
Interestingly, the final ensemble performance is significantly higher in the jointly trained one.
The result indicates that jointly training the sub-classifiers will encourage the sub-classifiers to have different roles to create synergy, and thus the final ensemble performance is higher than the independently trained one.
}

\begin{table}[!ht]
\centering
\resizebox{0.9\textwidth}{!}{%
\begin{tabular}{ l|c|c|c|c|c|c}
\hline
Architecture & Dataset & FC1 Acc & FC2 Acc & FC3 Acc & FC4 Acc & Averaging Acc
\\
\hline
\hline
ResNet-50~\cite{he2016deep} + SSM & ImageNet-1K & 65.24\% & 73.24\% & 75.09\% & 1.02\% & \textbf{76.68}\% \\
ResNet-50~\cite{he2016deep} + SSM-individual & ImageNet-1K & 75.60\% & 75.11\% & 76.18\% & 74.77\% & 75.38\% \\
\hline
\end{tabular}
}
\vspace{3mm}
\caption{\textbf{Results of ImageNet-1K classification according to two different training schemes.} \textit{SSM} is the result of training with the loss given to the averaged output; \textit{SSM-individual} is the result of training each output independently.}
\label{table:ablation1}
\end{table}


\subsubsection{Is SSM a new way of ensemble?}

\jc{
Ensemble is a simple technique to further boost performance by combining multiple models that have different random initializations.
The sub-classifiers of SSM may resemble the ensemble technique, and there may be concerns that SSM benefits from the ensemble-like effect and thus may not benefit from ensemble.
However, we argue that SSM is not simply an ensemble method, and we validate that SSM-augmented models can further benefit from ensemble.
}
\jc{
We train two ResNet-50 models and two ResNet-50 + SSM models with different initializations, and test if SSM can further benefit from ensembles.
The results are summarized in \tabref{table:ensemble}. The two ResNet-50 + SSM models' accuracies are 76.37\% and 76.68\%, and the ensembled accuracy is 78.04\%, which is 1.35\% higher. The improvement is a little less than the ResNet-50 ensemble, but it may be simply due to the performance saturation, and the 1.35\% is still a significant improvement by ensemble. Therefore, through this experiment we show that SSM-augmented models can further benefit from ensemble.
}

\begin{table}[!ht]
\centering
\resizebox{0.9\textwidth}{!}{%
\begin{tabular}{ l|c|c|c|c}
\hline
Architecture & Dataset & Epoch & Top-1 Accs & Ensembled Acc \\
\hline
\hline
ResNet-50~\cite{he2016deep} & ImageNet-1K & 90 & 75.60\%, 75.65\% & 77.25\% (+1.60\%)\\
ResNet-50~\cite{he2016deep} + SSM & ImageNet-1K & 90 & 76.37\%, 76.68\% & 78.04\% (\textbf{+1.35\%})\\
\hline
\end{tabular}
}
\vspace{3mm}
\caption{\textbf{Results of ensemble classification in ImageNet-1K.} All experiments are conducted in the same environment. We separately train the two models for two times each.}
\label{table:ensemble}
\end{table}

\subsubsection{Is the improvement simply due to parameter increases?}

\jc{
Finally, we show that the efficacy of SSM is not simply due to parameter increase.
To verify this, we further train two models with more parameters by adding more parallel classifiers.
As shown in \tabref{table:ab}, the base ResNet-50 has 25.55M parameters, and ResNet-50 + SSM has 28.58M parameter, so the parameter overhead is 3.03M.
One fully connected layer has 2.05M parameters, so we add one or two parallel fully connected layers to the baseline ResNet-50.
ResNet-50 (2FC) and ResNet-50 (3FC) are the new comparison methods that brings additional parameters in the classifier part, like SSM.
The result is summarized in \tabref{table:ab}.
A simple increase in parameters, like ResNet-50 (2FC) and (3FC), does not improves the performance much, but SSM does bring a significant improvement.
Therefore, we argue that the performance improvement is not simply due to parameter increase, but due to the feature exploiting characteristics of SSM.
}

\begin{table}[!ht]
\centering
\resizebox{0.9\textwidth}{!}{%
\begin{tabular}{ l|c|c|c|c}
\hline
Architecture & Dataset & Epoch & Top-1 Acc & Parameters \\

\hline
\hline
ResNet-50~\cite{he2016deep} & ImageNet-1K & 90 & 75.65\% & 25.55M\\
ResNet-50~\cite{he2016deep} + SSM & ImageNet-1K & 90 & \textbf{76.68\%} & 28.63M\\
ResNet-50~\cite{he2016deep} (2FC) & ImageNet-1K & 90 & 75.91\% & 27.60M\\
ResNet-50~\cite{he2016deep} (3FC) & ImageNet-1K & 90 & 75.67\% & 29.65M\\
\hline
\end{tabular}
}
\vspace{3mm}
\caption{\textbf{The result of the parameter increase in ImageNet-1K.} In all experiments, FC was added vertically, and all were ensembled using the averaging method.}
\label{table:ab}
\end{table}

\section{Qualitative analysis}
\label{sec:qualitative_analysis}

\jc{
The key intuition of SSM is to partially share the features among different sub-classifiers. As described in \sref{sec:fsm}, the first 1/4 channels are shared among all sub-classifiers, and the last 1/4 channels are used only by the last sub-classifier. The first 1/4 channels are the most frequently feed-forwarded and back-propagated, and are expected to contributes mostly to the final prediction of SSM. In short, our hypothesis is that the degree of sharing is positively correlated to the importance of the feature. Therefore, the first 1/4 channels are expected to contain the key features to classify among the target classes, and the last 1/4 channels are expected to contain additional features such as contextual information.
}

\begin{figure}[t!]
\begin{center}
    \includegraphics[height=18.22cm]{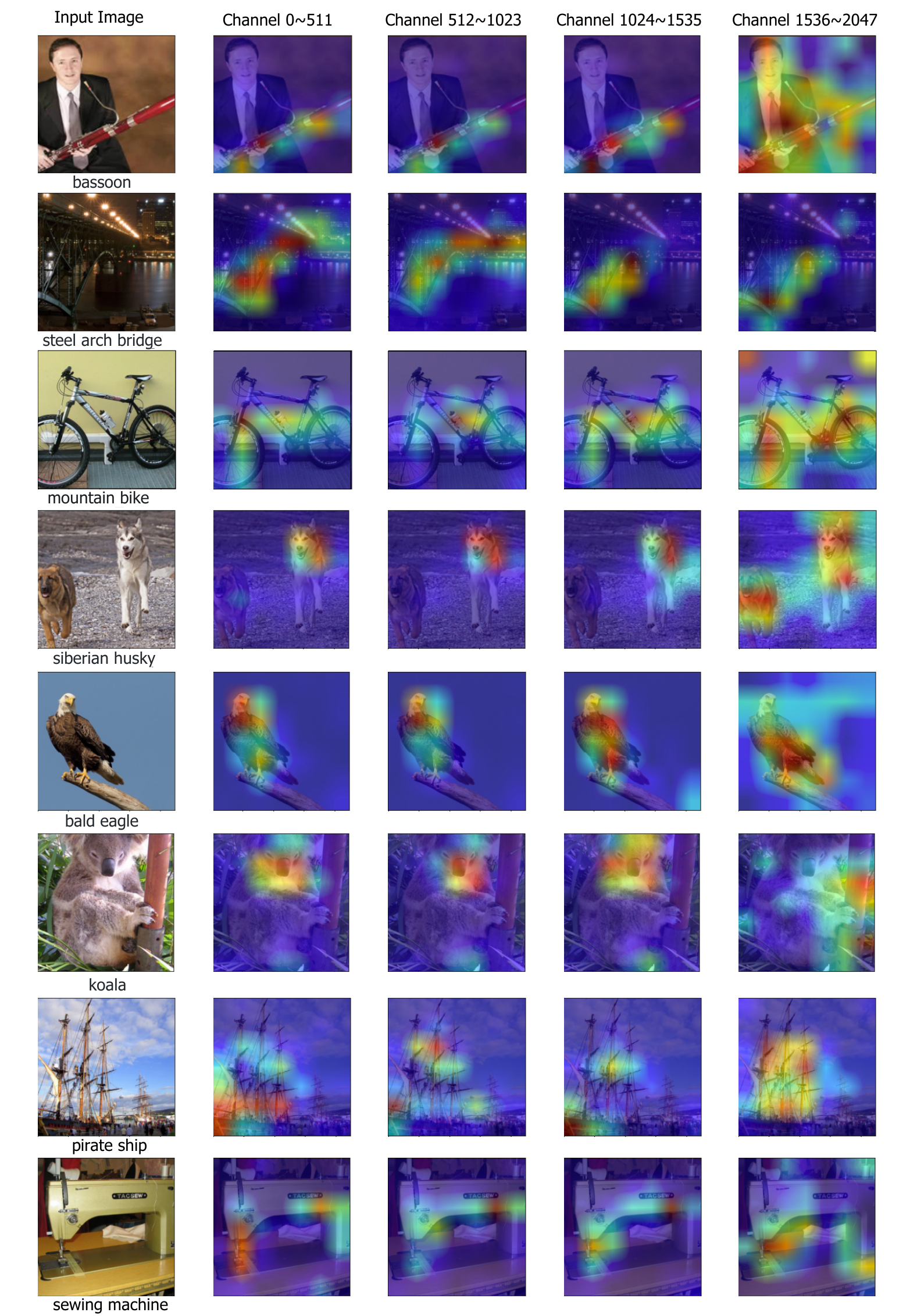}
    \caption{\textbf{Additional Grad-CAM visualization results.}}
    \label{figure:grad_cam2}
\end{center}
\end{figure}

\jc{
We qualitatively analyze the channels with the Grad-CAM visualization.
\figref{figure:grad_cam1} and \figref{figure:grad_cam2} shows input images from the validation set and the overlaid Grad-CAM heatmaps with respect to the ground-truth labels. ResNet-50+SSM is the visualization target. To analyze whether the feature splits have learned differently, the visualizations are generated for each 1/4 split of channels, instead of the full feature. The column `Channel 0\textasciitilde511' denotes the Grad-CAM of the first 1/4 channels with respect to the first sub-classifier. The column `Channel 512\textasciitilde1023' denotes the visualization of the second 1/4 channels w.r.t. the second sub-classifier, and so on. While the input to the second sub-classifier is the first 1/2 channels, we visualized the second 1/4 to explicitly compare the semantics learned in each 1/4 channels.
}

\jc{
The samples in \figref{figure:grad_cam1} and \figref{figure:grad_cam2} supports our intuition. The Grad-CAMs for the first sample in \figref{figure:grad_cam1} demonstrate that the first 1/4 channels focus on the ground-truth `guitar' location, and the last 1/4 channels focus on the corresponding context, which in this case is the guitar player. The two intermediate Grad-CAMs gradually changes from the key feature of the guitar to the corresponding context of the guitar player. The Grad-CAMs of the second sample in \figref{figure:grad_cam1} also show that the first 1/4 channels focus on the fish, and the last 1/4 channels focus on the corresponding context of the river. We demonstrate more samples in \figref{figure:grad_cam2}.
In summary, the Grad-CAM visualizations for each split channels show that the most shared channels focus on the target object, and the least shared channels focus on the corresponding context information.
}

\section{Discussion}
\label{sec:discussion}

We experimented with how much performance improvement would be achieved by selecting the best inferred output of each output and using it for prediction in the ResNet-50+SSM structure learned from the training set of ImageNet-1K. We have not yet developed an algorithm to select the optimal FC, so we have selected the optimal FC by ourselves, utilizing the label data of ImageNet-1K. The result was that if FC could be ideally chosen, an additional 6\% performance improvement could be seen with up to 82.9\% performance. Further research on this part is believed to be possible.

\section{Conclusion}
\label{sec:conclusion}

We propose Split-and-Share Module (SSM). SSM is a classifier that improves the performance of CNN networks. We apply the BatchNorm and ReLU to the shared features extracted from the feature extractor, limiting the number of commonly used and non-featured backprops, and have an effect of learning by placing weights on important features. Through this process, features learned according to importance have a classifier suitable for their capacity, and averaging multiple outputs from the classifier for use in training and testing. We verified SSM by applying CNNs of various structures in ImageNet-1K, and showed significant performance improvement in all the experiments. We also adopted Grad-CAM for qualitative analysis of SSM. Grad-CAM results showed qualitatively that our SSM could learn according to the importance of features as we intended. In addition, SSM divides features into four groups and learns features that are common or non-common features, and these characteristics are thought to be available in many areas of research.

\bibliography{egbib}
\end{document}